\documentclass[letterpaper, 10 pt, conference]{ieeeconf}  % Comment this line out if you need a4paper
\usepackage[T1]{fontenc}
\usepackage{amsmath}
\usepackage{algorithm}
\usepackage[noend]{algpseudocode}
\usepackage{graphicx} 
\usepackage{caption}
\usepackage{subcaption}
\usepackage{float}
\usepackage{hyperref}
\hypersetup{
    colorlinks=true,
    linkcolor=blue,
    filecolor=magenta,      
    urlcolor=cyan,
    }

\IEEEoverridecommandlockouts                              % This command is only needed if 
\title{\LARGE \bf
Learning multiple gaits of quadruped robot using \\hierarchical reinforcement learning
}

\author{Yunho Kim, Bukun Son, and Dongjun Lee
\thanks{Department of Mechanical Engineering, Seoul National University, Seoul, Republic of Korea. (e-mail: awesomericky@snu.ac.kr)}
}

\begin{document}

\maketitle
\thispagestyle{empty}
\pagestyle{empty}

%%%%%%%%%%%%%%%%%%%%%%%%%%%%%%%%%%%%%%%%%%%%%%%%%%%%%%%%%%%%%%%%%%%%%%%%%%%%%%%%
\begin{abstract}

There is a growing interest in learning a velocity command tracking controller of quadruped robot using reinforcement learning due to its robustness and scalability. However, a single policy, trained end-to-end, usually shows a single gait regardless of the command velocity. This could be a suboptimal solution considering the existence of optimal gait according to the velocity for quadruped animals \cite{horsegait, quadrupedgait}. In this work, we propose a hierarchical controller for quadruped robot that could generate multiple gaits (i.e. pace, trot, bound) while tracking velocity command. Our controller is composed of two policies, each working as a central pattern generator and local feedback controller, and trained with hierarchical reinforcement learning. Experiment results show 1) the existence of optimal gait for specific velocity range 2) the efficiency of our hierarchical controller compared to a controller composed of a single policy, which usually shows a single gait. Codes are publicly available \href {https://github.com/awesomericky/Multiple-gait-controller-for-quadruped-robot}{link}.

\end{abstract}

%%%%%%%%%%%%%%%%%%%%%%%%%%%%%%%%%%%%%%%%%%%%%%%%%%%%%%%%%%%%%%%%%%%%%%%%%%%%%%%%
\begin{figure*}[ht]
    \centering
    \includegraphics[width=\linewidth]{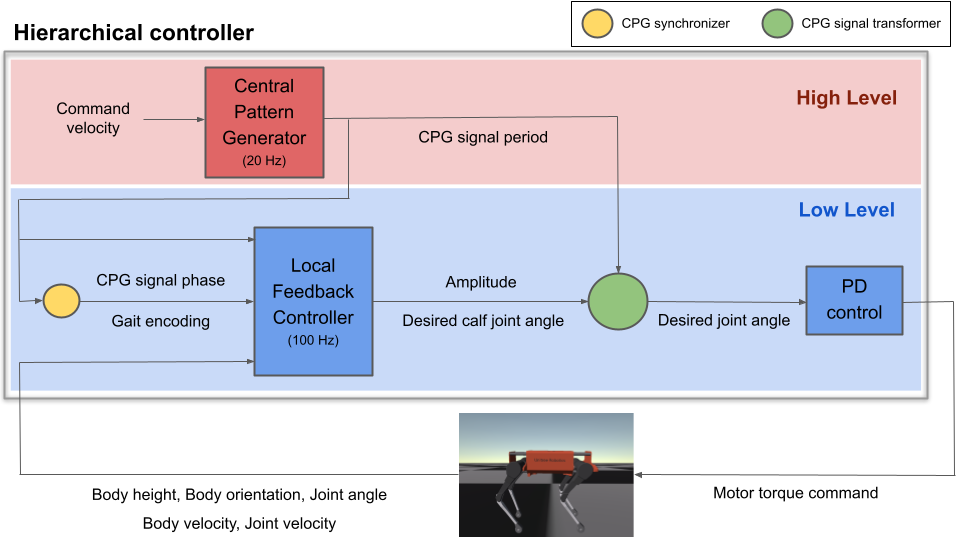}
    \caption{Proposed hierarchical controller for quadruped robot}
    % \vspace{-1.4em}
    \label{fig:controller}
\end{figure*}

\section{INTRODUCTION}

Designing a velocity command tracking controller of a quadruped robot is crucial for related applications. However, it is a complex problem due to rich contact with environment that is hard to model or predict. Reinforcement learning is an interesting solution for this due to its robustness and scalability which are shown in previous works \cite{legmotor, roughterrain}. Previous works used a simple framework composed of a single policy and trained it in end-to-end manner. However, this simple framework usually results a single gait regardless of the command velocity. 

According to the previous works done in biomechanics community, quadruped animals show different optimal gaits according to the velocity \cite{horsegait}. Similar results were also shown in simulated simple quadruped robot \cite{quadrupedgait}. Thus, the result of simple framework, which shows a single gait, could be a suboptimal solution. There could be advantages in perspective of energy usage and velocity tracking accuracy if we use multiple gaits. Thus, we present a novel hierarchical controller that could learn both single or multiple gaits. The details of the proposed controller will be explained in Section \ref{Method}

\section{RELATED WORKS}

\subsection{Locomotion control}

Designing a controller for agile locomotion, including gait pattern generation, is currently an active area of research in both simulation and real world robots. We could categorize previous approaches in three categories: model based analytic control, data driven control with reference motions, data driven control without reference motions.

First is model based analytic control. After mathematically modeling the dynamics and interactions of the robot, optimal solutions of the formulated problem could be used as control signal \cite{c1, c2, c3, c4, c5}. However, inaccurate dynamic model, complex optimization problem, stochastic environment, and online computation limits cause the approaches limited in robustness and scalability. Second is data driven control with reference motions. Reinforcement learning is often used in data driven control. Previous approaches designed simple reward to track reference motion and alleviated the difficulty of learning \cite{c6, c10}. The method was also successfully implemented in real world quadruped robot \cite{c7}. However, using reference motions and tracking reward signal to learn multiple behaviors could result suboptimal solution due to gradient conflicts and suboptimal reference motions. Some approaches alleviated the objective of strictly following reference motions using generative adversarial network \cite{c8, c9, c20}. Last is data driven control without reference motions. Designing specific reward signal enables learning desired behavior and often generates unexpected behavior \cite{legmotor, c12}. However, designing specific reward signal requires lots of trial and error and often difficult to learn multiple behaviors in a single policy.

\subsection{Gait generation}

Humans and animals show various gait patterns and smooth transitions depending on the environment and desired velocity. Researchers actively worked on generating and controlling multiple gaits considering the relationship between them \cite{c14, c19}. Erden et al. designed specific reward based on heuristics and learned adaptive gaits in limited environment \cite{c13}. Singla et al. manually time shifted 'walk' gait reference data to generate different gait data, and successfully restored other gaits using kinematic motion primitives \cite{c14}. Siekmann et el. proposed a simple reward signal to learn several bipedal gaits considering the periodicity of gait control signal \cite{c15}. However, previous approaches are quite limited in robustness and scalability because it is based on manual data generation process and limited heuristics. Furthermore, the methods didn't consider the relationship between the gait pattern and desired velocity, and rather designed the policy to take desired gait pattern as input signal.

\subsection{Quadruped gait}

Quadruped animals show various gaits (i.e. walk, pace, trot, bound etc). Previous researches experimentally showed the reason for gait selection and transition. Hoyt et al. discovered the relationship between gaits, velocity, and energy consumption in horse \cite{horsegait}. They found that each gait shows convex function of energy consumption according to the velocity and the minimum energy consumption region differs for each gait. Xi et al. found similar result with simulated 4 DOF quadruped robot \cite{quadrupedgait} and showed the necessity of using appropriate gaits for specific velocity range. According to these works, quadruped animals show walk gait in low velocity range, trot and pace gait in intermediate velocity range, bound and gallop gait in high velocity range.

\section{METHOD} \label{Method}

Our objective is to build a velocity command tracking controller that could show single or multiple gaits according to the command velocity. Previous approaches, which trained single policy in end-to-end manner were easily trapped to a single gait. We thought the reason for this was due to the distance between different gaits in joint angle domain, which is the output domain of the single policy. Different gaits correspond to different joint angle changes over time and they are often conflicting, which implies the large distance between different gaits in joint angle domain. Thus, the network could be easily trapped in a single gait region while not moving across different gait region. As mentioned in the previous work, using a single gait could result energy inefficient behavior and using multiple gaits appropriately could result more optimal behavior. To let the controller take advantage of diverse gaits and use each of them for appropriate velocity range, we propose a novel hierarchical controller (Fig \ref{fig:controller}). In this work, the controller was designed for 8 DOF (Degrees of Freedom) quadruped robot which has four thigh joints and four calf joints. However, it could also be applied to 12 DOF quadruped robot which also includes extra four hip joints.

\subsection{Framework}
Proposed controller is composed of two parts, the high level controller and low level controller. The high level controller plans and selects appropriate gait according to the command velocity. The role is similar to "Central Pattern Generator"(CPG) which is a biological neural circuits that produce simple rhythmic outputs for rhythmic behaviors like walking. The controller is composed of single policy and due to the similarity of role, we will call it as the central pattern generator. We parameterized CPG signal with simple sinusoidal function (Eq \ref{CPG}) and designed the central pattern generator to output the five parameters of sinusoidal function, $B$ for CPG signal period and $C_i$ $(i=0,1,2,3)$ for CPG signal phase where each corresponds to one leg, based on the command velocity (Eq \ref{high_policy}). We set the CPG signal period of four legs the same to facilitate learning natural rhythmic behavior. In this work, $C_i$ $(i=0,1,2,3)$, which are the CPG signal phase parameters, were manually selected and learning them from scratch will be left for future work.

\begin{equation} \label{CPG}
     CPG_{i}(t) = sin(Bt + C_i) ~\forall i, i\in\{0,1,2,3\}
\end{equation}
\begin{equation} \label{high_policy}
     \pi_{high}(v) = [ B, C_0, C_1, C_2, C_3 ]
\end{equation}

The low level controller enables forward walking or running by using the CPG signal while also considering feedback signal from the environment. The controller is composed of two parts, a single policy, which we will call as local feedback controller, and a PD controller. To directly connect CPG signal with gait pattern, we selected four thigh joints as CPG targets by assuming that they are more crucial for gait generation than calf joints. Then we parameterized the thigh joint angles using the CPG signal which will also result simple sinusoidal function with only difference in amplitude (Eq \ref{thigh_joint}). The amplitude $A$ roles as a domain transfer parameter from CPG signal domain to joint angle domain. We designed the local feedback controller to output five parameters, a domain transfer parameter $A$ and desired joint angles of four calf joints, based on the input composed of robot configuration and CPG feature (Eq \ref{low_policy}). Details of the input are written in Figure \ref{fig:controller}. Then eight desired joint angles are computed by transforming CPG signal to joint angle domain using the domain transfer parameter $A$ (Eq \ref{thigh_joint}, \ref{calf_joint}). The desired joint angles are passed to the PD controller to compute desired motor torques and are applied to the robot.

\begin{equation} \label{low_policy}
     \pi_{low}(s) = [ A, Calf_1, Calf_2, Calf_3, Calf_4 ]
\end{equation}
\begin{equation} \label{thigh_joint}
     Thigh_{i}(t) = A \times CPG_{i}(t) ~\forall i, i\in\{0,1,2,3\}
\end{equation}
\begin{equation} \label{calf_joint}
    Calf_{i}(t) = \pi_{low}(s)[i+1] ~\forall i, i\in\{0,1,2,3\}
\end{equation}

The low level controller is called more frequently then high level controller. In this work, the frequencies of high and low level controller were set to 20Hz, 100 Hz, however it could be changed freely based on the computation power of on-board computer of the robot. 

\begin{figure}[t]
    \centering
    \includegraphics[width=\linewidth]{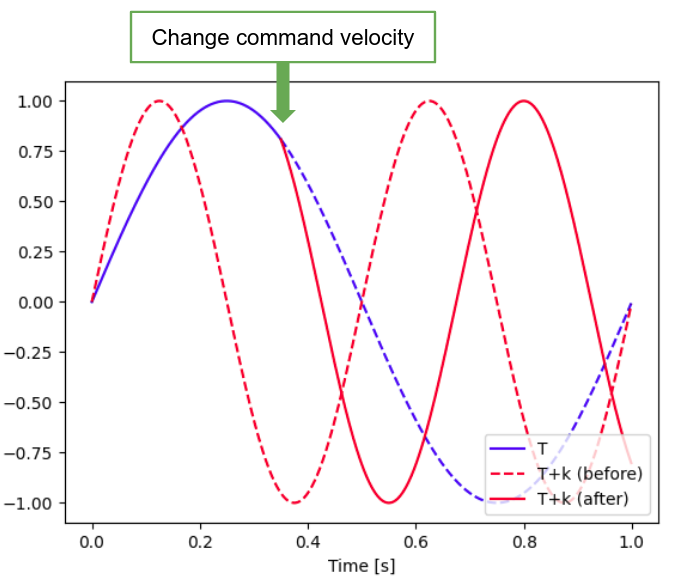}
    \caption{Effect of CPG signal synchronizer (T: current time, k: period of the high level controller)}
    \label{fig:CPG_synchronizer}
\end{figure}

In online scenario, there could be an abrupt change of CPG signal period due to the change of command velocity. If CPG signal shows discontinuity or sudden slope change, it could result unnatural behavior or even severe damages in real robot. Thus, we added a simple CPG signal synchronizer before the local feedback controller which works as Algorithm \ref{alg:synchronize}. The effect of CPG signal synchronizer is shown in Figure \ref{fig:CPG_synchronizer}.

The whole process of proposed hierarchical controller are summarized in Algorithm \ref{alg:hierarchy}.

\begin{algorithm*}[t]
\caption{CPG signal synchronization} \label{alg:synchronize}
\begin{algorithmic}[1]
\Procedure{Synchronize}{$B,$ $C,$ $period,$ $step,$ $\triangle t$}
\State $B_{old}\gets B$ 
\State $C_{old}\gets C$
\State $B_{new}\gets 2\pi/period$ 
\State $C_{new}\gets (B_{old} - B_{new}) \times step \times \triangle t + C_{old}$ 
\State \textbf{return} $B_{new}, C_{new}$ \Comment{synchronized CPG parameter}
\EndProcedure
\end{algorithmic}
\end{algorithm*}

\begin{algorithm*}[tb]
\caption{Hierarchical controller} \label{alg:hierarchy}
\begin{algorithmic}[1]
\State Set $\pi_{CPG}$ : Central pattern generator
\State Set $\pi_{local}$ : Local feedback controller
\State Initialize $B, C$
\State Set parameter $T_{CPG}$ : $\pi_{CPG}$ period
\State Set parameter $\triangle t$ : $\pi_{local}$ period
\State
\While {controller running}
\State Get command velocity
\If {$(step \times dt) \equiv 0 \mod T_{CPG}$}
\State $period\gets \pi_{CPG}(v)$
\State $B, C \gets SYNCHRONIZE(B,$ $C,$ $period,$ $step,$ $\triangle t)$ \Comment{CPG signal synchronization (\textbf{Algorithm 1})}
\State Generate CPG signal with $B, C$
\EndIf
\State Get current observation
\State Compute current CPG phase and gait encoding
\State $state \gets obsevation,$ $period,$ $CPG_{phase},$ $gait$ $encoding$
\State $A$, $Calf$ $joint$ $angle \gets \pi_{local}(state)$ 
\State $Thigh$ $joint$ $angle \gets$ $A \times CPG$ \Comment{CPG signal transformation}
\State Set $Thigh$ $and$ $Calf$ $joint$ $angle$ as target joint angle and compute motor torque with PD control
\State Execute the computed torque
\State $step \gets step + 1$
\EndWhile
\end{algorithmic}
\end{algorithm*}

\subsection{Training}

The controller was trained end-to-end using hierarchical reinforcement learning. For each policy, PPO algorithm, the state of the art model free reinforcement learning algorithm, was used \cite{PPO}. Similar cost terms with Hwangbo et al. and Lee et al. were used for realistic behavior with small changes \cite{legmotor, roughterrain} (Table \ref{tab:cost_table}). Each cost terms were weighted differently and integrated over time. The specific cost term equations and weighting parameters used for the experiment are summarized in APPENDIX (Eq \ref{cost:av}-\ref{cost:lp}, Table \ref{tab:hyperparameter}). Cost scale $k_c$ was gradually increased using $k_{c,j+1} \leftarrow (k_{c,j})^{k_d}$ $k_c, k_d \in(0,1)$ update rule and this let $k_c$ reach 1 gradually. Gradually increasing the cost scale avoids the robot from learning stopping behavior which is a suboptimal result. The update rule could be also thought of as part of curriculum learning, and had shown good result in previous works \cite{legmotor, bengio2009curriculum}.

\begin{table}[t]
    \centering
    \begin{tabular}{c|c}
    \hline
         angular velocity of the base cost & $c_1$ \\ \hline
         linear velocity of the base cost & $c_2$ \\ \hline
         torque cost & $c_3$ \\ \hline
         joint speed cost & $c_4$ \\ \hline
         foot vertical velocity cost & $c_5$ \\ \hline
         foot clearance cost & $c_6$ \\ \hline
         foot slip cost & $c_7$ \\ \hline
         orientation cost & $c_8$ \\ \hline
         smoothness cost & $c_9$ \\ \hline
         leg phase cost & $c_{10}$ \\ \hline
         \hline \\
         FINAL COST & c$_{final} = \sum_{i=1}^{10} w_i \cdot c_i$ \\ \\
         \hline
    \end{tabular}
    \caption{Cost terms used for training our hierarchical controller}
    \label{tab:cost_table}
\end{table}

\section{Experiment}

We deigned the experiment to validate the necessity of multiple gaits and the efficiency of proposed hierarchical controller. Particularly we aim to answer following three questions.
\begin{enumerate}
    \item Does there exist optimal gait for 8 DOF quadruped robot according to velocity?
    \item Can the proposed hierarchical controller work as a multiple gait controller?
    \item How does the multiple gait controller perform compared to a single gait controller?
\end{enumerate}

\subsection{Setup}

We experimented our hierarchical controller in RAISIM simulator \cite{raisim}. Laikago quadruped robot model from Unitree Robotics was used for training. Although laikago has 12 DOF, we fixed four hip joints, resulting 8 DOF, and focused on learning just forward moving behavior. However, as the proposed controller has no limited constraint for number of joints, expanding the work by including four hip joints will be left for future work. For computation, single GPU was enough due to the light weight network model. 

Hyperparameters and network architectures are written in APPENDIX Table \ref{tab:hyperparameter}. Same hyperparameters and architectures were used for the whole experiments.

\subsection{Comparison between different gaits}

To check the existence of optimal gait according to the velocity, we learned different gaits separately using our controller. Then we defined optimality with velocity tracking error and energy consumption and measured the values for each gait. Trot, pace, and bound gaits were chosen to be learned because they are the three most typical gaits of quadruped animals. CPG signal phase parameter, $C_i$ in Equation \ref{CPG}, \ref{high_policy}, were manually fixed with values shown in Table \ref{tab:gaitphase}.

\begin{table} [t]
    \centering
    \begin{tabular}{c|c}
         & phase \\ \hline
        Trot & $[\pi, 0, 0, \pi]$ \\ \hline
        Pace & $[\pi, 0, \pi, 0]$ \\ \hline
        Bound & $[\pi, \pi, 0, 0]$ \\
    \end{tabular}
    \caption{CPG signal phase for each gait. The order of leg phases is [FR, FL, RR, RL]}
    \label{tab:gaitphase}
    \vspace{-5mm}
\end{table}

\begin{figure*}[ht]
    \centering
    \includegraphics[width=\linewidth]{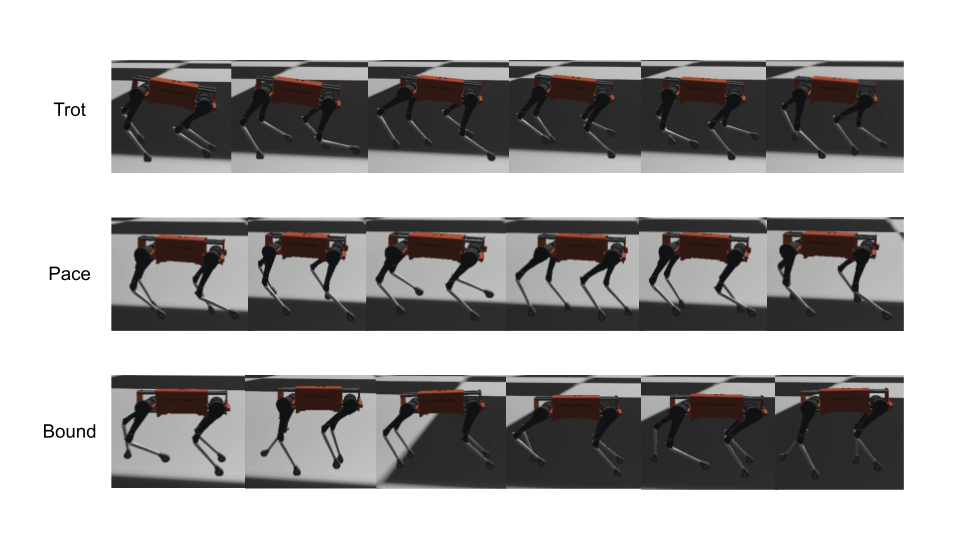}
    \caption{Single gait learned using our hierarchical controller}
    \label{fig:exp1_gait}
\end{figure*}

\begin{figure*}[ht]
    \centering
    \includegraphics[width=\linewidth]{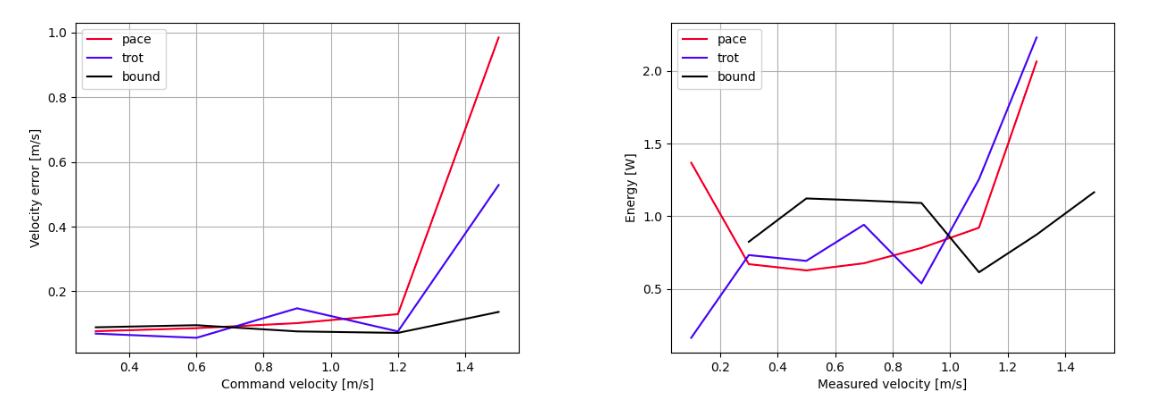}
    \caption{Velocity tracking error (left) and Energy consumption (right) for each gait}
    \label{fig:exp1_performance}
\end{figure*}

Every gaits were almost successfully learned as shown in Figure \ref{fig:exp1_gait}. Bound gait showed incomplete result due to contact with the environment which caused CPG signal phase shift  of RR and RF thigh joints. For each learned gaits, we measured the mean velocity tracking error and energy consumption. 0.3, 0.6, 0.9, 1.2, 1.5 $m/s$ command velocity were given as input, which are uniformly selected velocities in the trained velocity range. For the energy consumption, we divide the whole measured velocity range by 0.2 and clustered using 0.1, 0.3, 0.5, 0.7, 0.9, 1.1, 1.3, 1.5 $m/s$ as median velocity for each equalized range. Then the mean energy consumption for each range were computed. The results are shown in Figure \ref{fig:exp1_performance}. Pace and trot showed smaller energy usage in low velocity range while bigger in high velocity range compared to bound. Pace and trot also failed to learn high velocity tracking while bound succeeded. These results are similar to previous works with horse and simulated 4 DOF quadruped robot \cite{horsegait, quadrupedgait}. From the result we could conclude that pace and trot are optimal for intermediate velocity range, while bounds are optimal for high velocity range.

\begin{figure*} [ht]
    \centering
    \includegraphics[width=\linewidth]{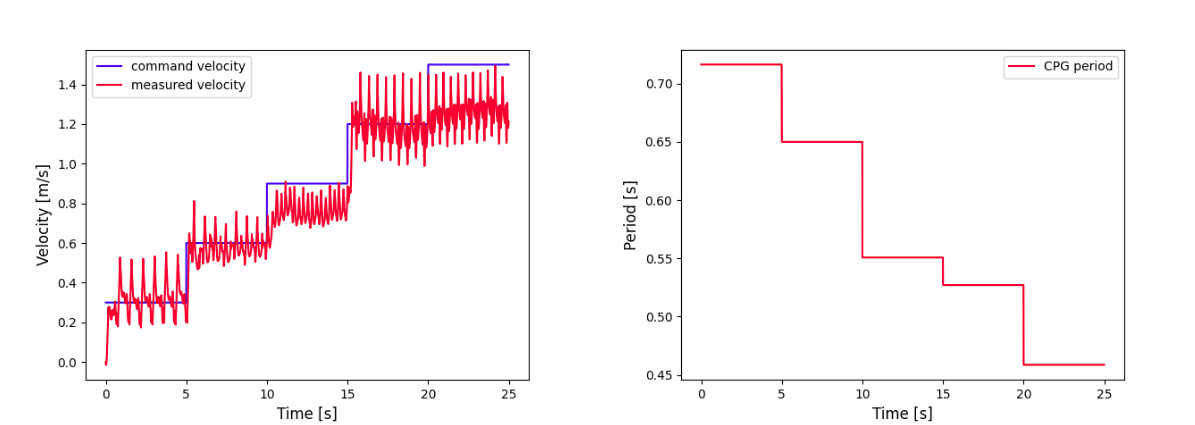}
    \caption{Velocity tracking result (left) and CPG signal period (right) of multiple gait controller}
    \label{fig:exp2_velocity}
\end{figure*}

\begin{figure*}[ht]
    \centering
    \includegraphics[width=\linewidth]{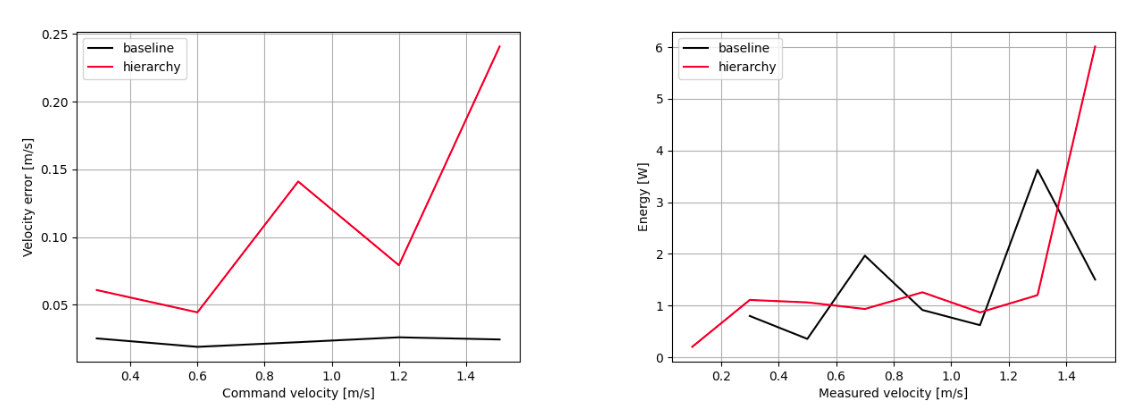}
    \caption{Velocity tracking error (left) and Energy consumption (right) of multiple gait controller and baseline controller}
    \vspace{-1.4em}
    \label{fig:exp2_performance}
\end{figure*}

\subsection{Performance of multiple gait controller}

Based on the performance of different gaits examined in previous experiment, we learned a multiple gait controller using our framework by defining a function that determines CPG signal phase feature according to the velocity as Equation \ref{exp2_CPG_phase_function}. Then we used it for the high level controller output as Equation \ref{high_policy_exp2}. The purpose of the defined function was to generate trot, pace, bound in $[0, 0.5]$, $[0.5, 1]$, $[1, 1.5]$ velocity range.

\begin{equation} \label{exp2_CPG_phase_function}
    f(v) = 
    \begin{cases}
    [\pi, 0, 0, \pi] & 0 < v \leq 0.5 \\
    [\pi, 0, \pi, 0] & 0.5 < v \leq 1 \\
    [\pi, \pi, 0, 0] & 1 < v \leq 1.5
    \end{cases}
\end{equation}

\begin{equation} \label{high_policy_exp2}
     \pi_{high}(v) = [ B, f(v) ]
\end{equation}

The velocity tracking result and CPG signal period of the learned multiple gait controller are shown in Figure \ref{fig:exp2_velocity}. By using appropriate gait for each velocity range, we nearly tracked the command velocity. Furthermore, the CPG signal period, which is the output of high level controller, decreased as the command velocity increased. From the result, we could realize that our proposed hierarchical controller can learn multiple gaits by taking the advantage of both high and low level controller each working as a central pattern generator and local feedback controller. Furthermore, each polices, composing the controller, converged to a reasonable outcome.

We compared the performance of multiple gait controller with a simple framework composed of a single policy. The baseline was trained with PPO using the same reward signal shown in Table \ref{tab:cost_table}. However, as we couldn't define swing and stance phase in the baseline method, we set the leg phase cost coefficient to zero. The cost coefficients for other costs were set to the same value as in Table \ref{tab:hyperparameter}. The trained baseline method converged to a single gait as expected and showed unnatural behavior. 

We compared the performance of multiple gait controller and the baseline by measuring velocity tracking error and energy consumption, same as previous experiment. The results are shown in Figure \ref{fig:exp2_performance}. Multiple gait controller showed energy efficiency in some velocity region compared to the baseline controller. However, the velocity tracking error of multiple gait controller was much higher compared to the baseline. This is due to abrupt change of CPG signal which corresponds to abrupt change of gait. Algorithm \ref{alg:synchronize} could handle abrupt change of CPG period, but not the abrupt change of CPG phase which occurred due to gait transition. These discontinuity of CPG signal results degraded performance of tracking the command velocity. We will expand Algorithm \ref{alg:synchronize} to handle difference in CPG phase for smooth gait transition as a future work.

\section{CONCLUSIONS}

We propose a novel hierarchical controller where high and low level controller collaborate to generate single or multiple gaits. The high level controller is composed of central pattern generator which plans and selects gaits. The low level controller is composed of local feedback controller and PD controller which enable walking or running by using the CPG signal and the environment feedback. In experiment, we showed the existence of optimal gait for 8 DOF quadruped robot according to the velocity. Furthermore, we empirically showed that our proposed hierarchical controller could learn multiple gaits that could be effective compared to single gait controller. However, the efficiency of multiple gait controller should be more thoroughly studied using various baselines including analytic model based controller. Also the performance of multiple gait controller should be improved by considering smooth gait transition and CPG signal phase shift problem. We will leave further improvements as future work.

\addtolength{\textheight}{-2cm}   % This command serves to balance the column lengths

%%%%%%%%%%%%%%%%%%%%%%%%%%%%%%%%%%%%%%%%%%%%%%%%%%%%%%%%%%%%%%%%%%%%%%%%%%%%%%%%

%%%%%%%%%%%%%%%%%%%%%%%%%%%%%%%%%%%%%%%%%%%%%%%%%%%%%%%%%%%%%%%%%%%%%%%%%%%%%%%%

%%%%%%%%%%%%%%%%%%%%%%%%%%%%%%%%%%%%%%%%%%%%%%%%%%%%%%%%%%%%%%%%%%%%%%%%%%%%%%%%
% \clearpage
\section*{APPENDIX}

\subsection{Cost terms}

Below are the cost terms used for training the proposed hierarchical controller. To control command tracking error, which is the main objective of the controller, two different logistic kernels were used for angular velocity and linear velocity cost (Eq \ref{costkernel:av}, \ref{costkernel:lv}).
\begin{equation} \label{costkernel:av}
    K_{angular}(x) = - \frac{1}{e^{10x} + 2 + e^{-10x}}
\end{equation}
\begin{equation} \label{costkernel:lv}
    K_{linear}(x) = - \frac{1}{e^x + 2 + e^{-x}} - \frac{1}{e^{10x} + 2 + e^{-10x}}
\end{equation}

\noindent \textbf{Notation}\\
\makebox[1.2cm]{$k_c$} cost scale.\\
\makebox[1.2cm]{$k_d$} curriculum factor.\\
\makebox[1.2cm]{${{v}}^C_{AB}$} linear velocity of $B$ respect to $A$ expressed in $C$\\
\makebox[1.2cm]{$\omega$} angular velocity\\
\makebox[1.2cm]{$\hat{\cdot}$} desired quantity\\
\makebox[1.2cm]{$\tau$} joint torque\\
\makebox[1.2cm]{$\phi$} angular quantity\\
\makebox[1.2cm]{$v_{ft}$} tangential velocity of a foot (x, y components)\\
\makebox[1.2cm]{$v_{fz}$} vertical velocity of a foot (z components) \\
\makebox[1.2cm]{$p_f$} linear position of a foot\\
\makebox[1.2cm]{$g_i$} contact function of $i_{th}$ foot \\
\makebox[1.2cm]{} (0: not in contact, 1: in contact)\\
\makebox[1.2cm]{$G_i$} Leg phase function of $i_{th}$ foot \\
\makebox[1.2cm]{} (0: swing phase, 1: stance phase)\\
\makebox[1.2cm]{$a_t$} action at $t$ step \\
\makebox[1.2cm]{$\lvert \cdot \rvert$} cardinality of a set or $l_1$ norm\\
\makebox[1.2cm]{$\lvert\lvert \cdot \rvert\rvert$} $l_2$ norm\\

\vspace{0.15cm}
\noindent \textbf{angular velocity of the base cost}
\begin{equation} \label{cost:av}
    c_{1} = K_{angular}(k_c ||{\omega}^I_{IB}-\hat{{\omega}}^I_{IB}||^{2})
\end{equation}
\vspace{0.15cm}
\textbf{linear velocity of the base cost}
\begin{equation} \label{cost:lv}
    c_{2} = K_{linear}(|{v}^I_{IB}-\hat{{v}}^I_{IB}|)
\end{equation}
\vspace{0.15cm}
\textbf{torque cost}
\begin{equation} \label{cost:torque}
    c_{3} = k_c ||\tau||
\end{equation}
\vspace{0.15cm}
\textbf{joint speed cost}
\begin{equation} \label{cost:js}
    c_{4} = k_c \lvert \dot{{\phi}}^{i} \rvert ^2 \quad \forall i \in \{1,2...,8\}
\end{equation}
\vspace{0.15cm}
\textbf{foot vertical velocity cost}
\begin{equation} \label{cost:fvv}
    c_{5} = k_c |v_{fz,i}|^{2}, ~\forall i, i\in\{0,1,2,3\}
\end{equation}
\vspace{0.15cm}
\textbf{foot clearance cost} ($\hat{p}_{f,i,z} = 0.07 ~\textrm{m}$)
\begin{equation} \label{cost:fc}
\begin{split}
    c_{6} = k_c (max(0, \hat{p}_{f,i,z} - p_{f,i,z}))^2||v_{ft,i}||, \\ ~\forall i, g_i = 0, i\in\{0,1,2,3\}
\end{split}
\end{equation}
\vspace{0.15cm}
\textbf{foot slip cost}
\begin{equation} \label{cost:fs}
    c_{7} = k_c ||v_{ft,i}||, ~\forall i, g_i = 1, i\in\{0,1,2,3\}
\end{equation}
\vspace{0.15cm}
\textbf{orientation cost}
\begin{equation} \label{cost:oc}
    c_{8} = k_c ||[0,0,1]^T - \phi_{g}||
\end{equation}
\vspace{0.15cm}
\textbf{smoothness cost}
\begin{equation} \label{cost:sc}
    c_{9} = k_c ||a_{t-1} - a_{t}||
\end{equation}
\vspace{0.15cm}
\textbf{leg phase cost}
\begin{equation} \label{cost:lp}
    c_{10} = \frac{1}{4} (g_iG_i + (1 - g_i)(1 - G_i)) ~\forall i, i \in \{0,1,2,3\}
\end{equation}

\subsection{Hyperparameters and Network architecture}

\begin{table}[H]
\centering
\begin{tabular}{|p{0.5 \textwidth}|}
\hline
\\
\noindent \textbf{Hyperparameter}\\
\makebox[1.2cm]{$v$} Uniform(0.1, 1.5) [m/s] \\
\makebox[1.2cm]{$w_1$} 120. \\
\makebox[1.2cm]{$w_2$} 500. \\
\makebox[1.2cm]{$w_3$} 0.5\\
\makebox[1.2cm]{$w_4$} 0.02 \\
\makebox[1.2cm]{$w_5$} 1.0 \\
\makebox[1.2cm]{$w_6$} $1.5 \times 10^4$ \\
\makebox[1.2cm]{$w_7$} 200. \\
\makebox[1.2cm]{$w_8$} 100. \\
\makebox[1.2cm]{$w_9$} 0.5 \\
\makebox[1.2cm]{$w_{10}$} 300. \\
\makebox[1.2cm]{$k_{c,0}$} 0.3 \\
\makebox[1.2cm]{$k_d$} 0.999 \\
\\
\noindent \textbf{Network architecture}\\
\begin{tabular}{ l  l }
  hidden units (high) & [128] \\
  hidden units (low) & [128, 128] \\
  activation (high) & LeakyReLU \\
  activation (low) & LeakyReLU \\
\end{tabular} \\
\\
\hline
\end{tabular}
\caption{Hyperparameters and network architecture} 
\label{tab:hyperparameter}
\end{table}

\subsection{Additional plots for learned single gait}
\begin{figure}[H]
    \centering
    \includegraphics[width=\linewidth]{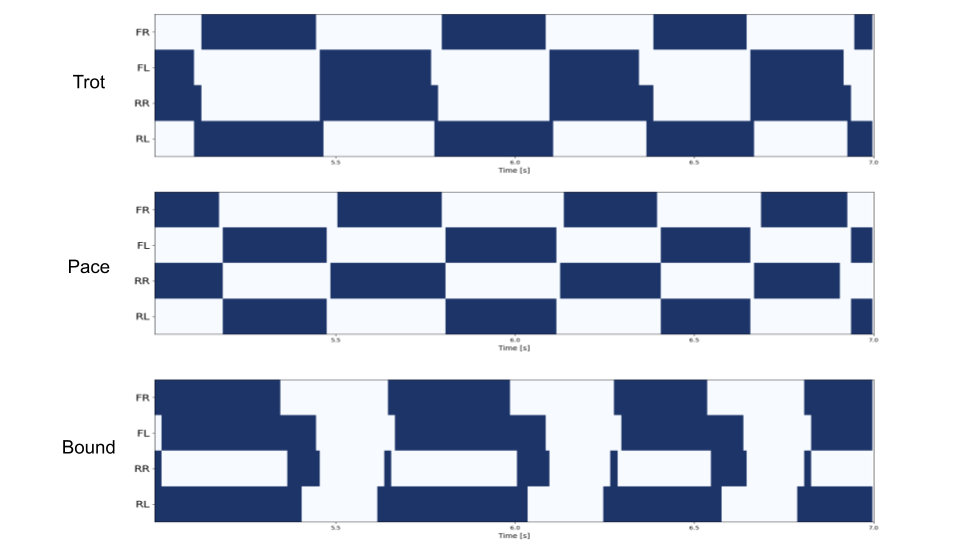}
    \caption{Contact plot for each gait}
    \label{fig:appendix_contact}
\end{figure}
\begin{figure*}
    \centering
    \begin{subfigure}[b]{\textwidth}
        \centering
        \includegraphics[width=\textwidth]{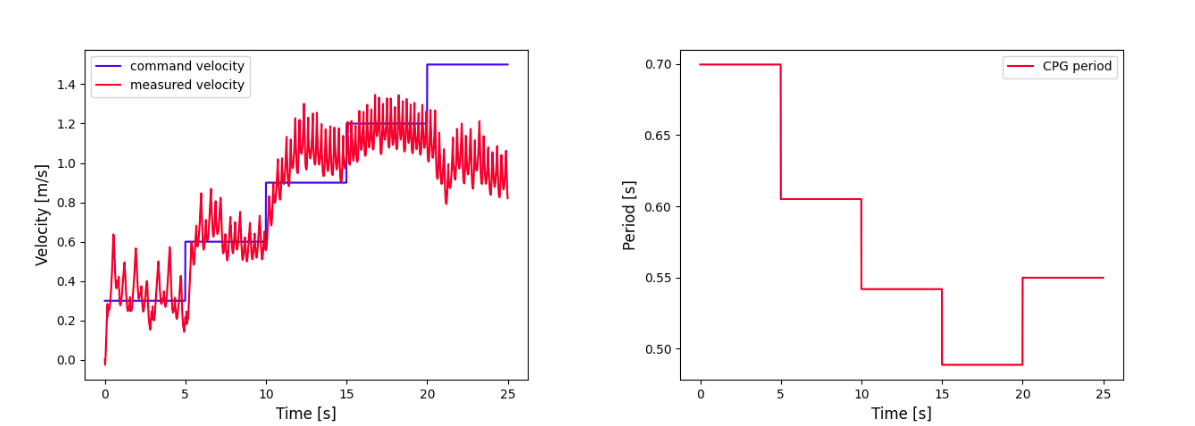}
        \caption{Trot}
        \label{fig:appendix_trot}
    \end{subfigure}
    \hfill
    \begin{subfigure}[b]{\textwidth}
        \centering
        \includegraphics[width=\textwidth]{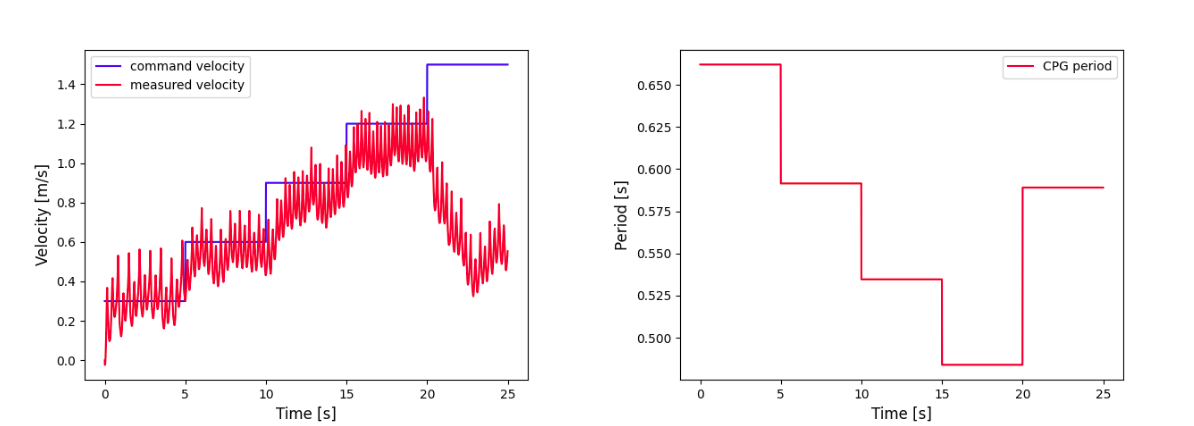}
        \caption{Pace}
        \label{fig:appendix_pace}
    \end{subfigure}
    \hfill
    \begin{subfigure}[b]{\textwidth}
        \centering
        \includegraphics[width=\textwidth]{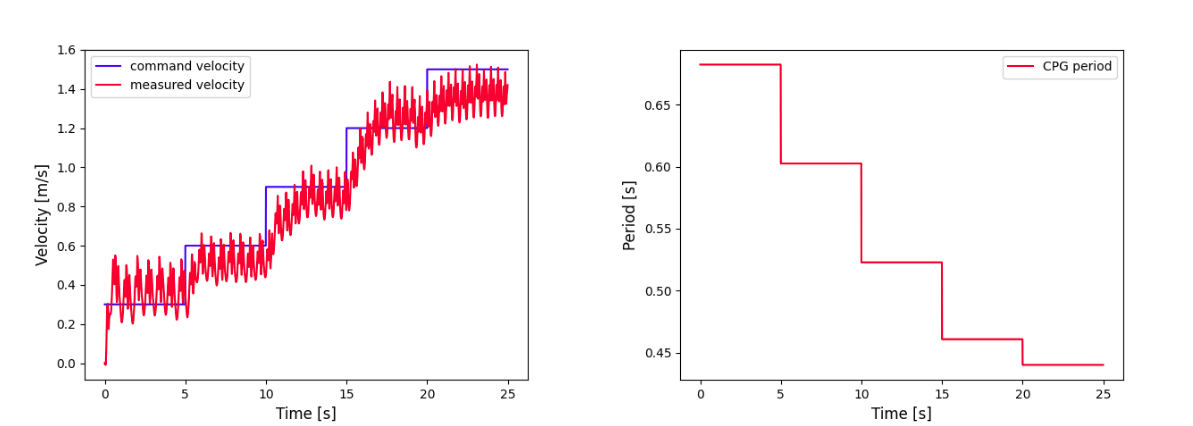}
        \caption{Bound}
        \label{fig:appendix_bound}
    \end{subfigure}
    \caption{Velocity tracking result (left) and CPG signal period (right) for each gait}
    \label{fig:appendix_single_gait}
\end{figure*}

%%%%%%%%%%%%%%%%%%%%%%%%%%%%%%%%%%%%%%%%%%%%%%%%%%%%%%%%%%%%%%%%%%%%%%%%%%%%%%%%

\end{document}